\documentclass{article}

\usepackage[preprint]{neurips_2024}

\usepackage{subcaption}
\usepackage{graphicx}
\usepackage[utf8]{inputenc} % allow utf-8 input
\usepackage[T1]{fontenc}    % use 8-bit T1 fonts
\usepackage{hyperref}       % hyperlinks
\usepackage{url}            % simple URL typesetting
\usepackage{booktabs}       % professional-quality tables
\usepackage{float}
\usepackage{amsfonts}       % blackboard math symbols
\usepackage{nicefrac}       % compact symbols for 1/2, etc.
\usepackage{microtype}      % microtypography
\usepackage{xcolor}         % colors

\title{On the Dynamics of Multi-Agent LLM Communities Driven by Value Diversity}

\author{
  Muhua Huang$^{1}$, Qinlin Zhao$^{2}$, Xiaoyuan Yi$^{2}$\thanks{Corresponding author.}\ ,~~Xing Xie$^{2}$ \\
  $^{1}$Stanford University, $^{2}$Microsoft Research Asia \\
  \texttt{muhua@stanford.edu,\{v-zhaoqinlin,xiaoyuanyi,xing.xie\}@microsoft.com}
}

\begin{document}

\maketitle

\begin{abstract}
As Large Language Models (LLM) based multi-agent systems become increasingly prevalent, the collective behaviors, \textit{e.g.}, collective intelligence, of such artificial communities have drawn growing attention. This work aims to answer a fundamental question: \emph{How does diversity of values shape the collective behavior of AI communities?} Using naturalistic value elicitation grounded in the prevalent Schwartz's Theory of Basic Human Values, we constructed multi-agent simulations where communities with varying numbers of agents engaged in open-ended interactions and constitution formation. The results show that value diversity enhances value stability, fosters emergent behaviors, and brings more creative principles developed by the agents themselves without external guidance. However, these effects also show diminishing returns: extreme heterogeneity induces instability. This work positions value diversity as a new axis of future AI capability, bridging AI ability and sociological studies of institutional emergence.
\end{abstract}

\section{Introduction}
Recent advances in AI have been driven by scaling up neural networks\citep{kaplan2020scalinglawsneurallanguage,hoffmann2022training,gao2023scaling}, yielding powerful Large Language Models (LLMs)~\citep{bubeck2023agisparks,openai2024gpt4,guo2025deepseek_r1}. However, simply making models larger is facing diminishing returns in performance~\citep{caballero2023brokenneuralscalinglaws,chen2025revisiting}. This plateau has motivated exploration of new approaches to further push the boundaries of LLM capability.

In parallel, research on humans has demonstrated the potential of \emph{collective intelligence} (CI), \textit{a.k.a}, the $c$-factor: groups can exhibit a measurable general intelligence ability to perform well across a wide range of tasks~\citep{woolley2010evidence}. Such a $c$-factor is not strongly tied to the average or maximum individual intelligence of members but instead correlates with group processes~\citep{riedl2021quantifying}. This has sparked a range of research on the collective intelligence of LLMs~\citep{ferreira2024organizing,burton2024large,talebirad2025wisdom}. Notably, prior work shows that diversity of perspectives can enhance group problem-solving, \textit{e.g.}, a team of diverse problem solvers can outperform that of uniformly high-ability solvers~\citep{hong2004groups}. 

Drawing inspiration from these insights, before moving on to the broader CI problem, in this work, we begin with a pilot study that investigates \emph{how value diversity shapes the collective behavior of AI agent communities}. Given that values are a core aspect of human diversity~\citep{inglehart2005modernization,sagiv2022personal} and are tightly linked to AI safety~\citep{bai2022traininghelpfulharmlessassistant} and user preferences~\citep{rafailov2023direct}, we focus on the value-driven dynamics of LLMs, which may provide a broader repertoire of heuristics, enabling groups to escape local optima.

This preliminary work is situated at the intersection of several research threads. First, it extends the study of collective behaviors from humans~\citep{woolley2010evidence,riedl2021quantifying} into the domain of AI-only groups. We ask whether similar principles, \textit{e.g.}, balanced participation, effective collaboration, apply when the ``agents'' are LLMs. Second, it draws on multi-agent social simulation using LLMs. Recent studies have indicated that LLM agents can engage in rich interactions, forming relationships/norms and exhibiting believable social behaviors in open-ended environments without a predefined objective~\citep{park2023generativeagentsinteractivesimulacra,zhang2025evolvingcollectivecognitionhumanagent}. This suggests emergent group dynamics can arise from AI-agent interactions alone. Third, our work connects to AI safety and alignment~\citep{bai2022traininghelpfulharmlessassistant}. Prior approaches tried to align a single model with a set of principles or ``constitution'' created by developers~\citep{bai2022constitutional} or derived from public input~\citep{Huang_2024}, to guide its behavior. We explore a complementary angle: rather than hand-coding a constitution, can a community of value-diverse AI agents self-organize their own norms and balances?

In this paper, we present an in-depth simulation study of LLM-based agent communities varying in value diversity. Groups of 4, 10, and 30 agents were instantiated with personas derived from Schwartz's Theory of Basic Human Values~\citep{schwartz2012overview}, ensuring a range of motivational value orientations. Crucially, instead of assigning values abstractly, we developed a naturalistic persona elicitation method: each agent ``discovers'' its values by solving a series of ethical dilemmas, mimicking how humans form values through life experiences~\citep{bai2025irotehumanliketraitselicitation}. This approach grounds the agents' personas in concrete decisions, yielding more realistic, nuanced value profiles. We then observe the agents through a two-stage interaction protocol, including (1) free-form open discussion, (2) collaborative rule formation to evaluate their emergent collective behavior. 

We ask: How does value diversity influence the group's conversational dynamics? We include control groups with no explicit value priming to isolate the effects of our value-based diversity interventions. Our findings reveal that \emph{value diversity} can significantly boost collective behaviors.
\section{Related Work}
\paragraph{Collective Intelligence}
The concept of a collective intelligence factor originates from human group research. \cite{woolley2010evidence} demonstrated that groups show a consistent ability across tasks, analogous to individual general intelligence. They measured CI by having groups complete a battery of diverse tasks (brainstorming, puzzle solving, moral reasoning, etc.) and found a single factor explained much of the variance in group performance across tasks. Subsequent studies reinforced the idea of a $c$-factor. \cite{riedl2021quantifying} aggregated data from 22 studies (over 1,300 groups) and found that process matters: groups with better collaboration processes, \textit{e.g.}, taking turns, listening, coordinating, achieved higher CI, even above what one would expect from individual talents. These works inform our operationalization of CI in AI agent groups, where we treat balanced participation (no single agent dominating conversation) as one indicator of a healthy collective process. Diversity's role in collective intelligence has been widely studied in human contexts. Diversity can be along dimensions of knowledge, cognitive styles, values, demographics, etc. \cite{hong2004groups} classic computational study showed that diversity trumps ability in many problem-solving scenarios: a heterogeneous group of solvers found better solutions than a homogeneous group of top-performing individuals, primarily because diverse agents explored solution space more broadly.

\paragraph{Multi-Agent LLM Simulations}
Using LLMs as agents in interactive simulations is a nascent but rapidly growing area. Aside from the aforementioned Generative Agents work \citep{park2023generativeagentsinteractivesimulacra, zhang2025evolvingcollectivecognitionhumanagent} where agents lived in a sandbox environment, other studies have had LLM agents play roles in games, negotiations, and collaborations. Agents have also been used to simulate surveys or social scenarios \citep{huang2025designingaiagentspersonalitiespsychometric} to study how different ``personality'' prompts might respond. A key takeaway is that LLM agents can display consistent persona traits and remember interactions to a degree, enabling multi-round emergent dynamics. However, many prior simulations assign roles or goals explicitly. In contrast, we give agents high-level values but no specific task or explicit role, to observe truly open-ended social emergence. Our work also examines emergent governance, which relates to studies on norm emergence in multi-agent systems \citep{lai2024evolving,cordova2024systematicreviewnormemergence, sarkar2024normative}. Researchers have shown that even simple agents can evolve cooperation strategies or social norms. With LLM agents, there is an opportunity to see richer normative behaviors. 

\paragraph{AI Alignment and Value System}
Current AI alignment approaches typically rely on simplified frameworks that operationalize human values into trainable objectives. The widely-adopted HHH framework (Helpful, Harmless, Honest)~\citep{askell2021generallanguageassistantlaboratory} attempts to capture core human values, providing a foundation for initial alignment. However, this reductive approach struggles with complex value trade-offs that arise in real-world scenarios where helpfulness, harmlessness, and honesty may conflict, failing to capture the complex motivations, long-term planning, and social reasoning that characterize emerging autonomous systems. To address these limitations, researchers have turned to more comprehensive value frameworks. Schwartz's Theory of Basic Human Values identifies ten universal values arranged in a circular structure where adjacent values are compatible and opposing values create tension~\citep{schwartz2012overview}. Recent work has begun mapping LLM behavior onto this value space to evaluate and guide LLMs~\citep{yao2023value,duan2025adaem}. This approach moves beyond binary alignment to consider how AI systems might embody different value profiles for different contexts and navigate the complex trade-offs that agentic systems face.

Our study leverages Schwartz's framework to initialize agent diversity in a principled way, ensuring a broad coverage of the value space. Rather than pre-programming values, we use realistic ethical scenarios to elicit each agent's values (as described below), an approach we believe yields more authentic and multi-dimensional personas. On the alignment front, our simulation speaks to collective alignment mechanisms. In our multi-agent setting, we observe whether agents can collectively draft their own rules of conduct, effectively creating a constitution organically. This resonates with ideas of decentralized AI governance, where no single authority dictates the rules, but rather rules emerge from many agents' inputs. Such self-constraints among AI could be crucial in open-world deployments where multiple AI systems interact without constant human oversight. By studying this in a controlled environment, we hope to inform how value-diverse AI might coordinate and check one another, potentially leading to safer outcomes. Next, we detail our methods, including how we instantiate agent values and measure CI across interaction stages.
\section{Methods}
We conducted an agent-based simulation experiment with LLM-driven agents. The experiment manipulated three key independent variables: (1) Group Size, (2) Value Composition, and (3) Value Complexity of the agent community. Each agent in every group was implemented by the same base language model (LLaMA-3.1-70B) to hold language capabilities constant, but was given a unique persona profile influencing its behavior. Figure 1 illustrates our overall approach, including the persona elicitation pipeline and the cognitive architecture of the agents (internal state and memory modules). The simulation proceeded in two stages, free-form interaction and governance emergence, with detailed logging and periodic assessments to track the collective state. Below, we describe the agent design, experimental conditions, simulation procedure, and measurement of collective behavior.

\subsection{Agent Personas and Value Elicitation}
Rather than assigning agents superficial traits or fixed value labels, we used a naturalistic value formation approach to create nuanced personas. We drew on Schwartz's Theory of Basic Human Values, which identifies 10 universal value types grouped into four higher-order categories \citep{schwartz2012overview}. The ten basic values are: Self-Direction, Stimulation (Openness to Change); Achievement, Power, Hedonism (Self-Enhancement); Security, Conformity, Tradition (Conservation); and Benevolence, Universalism (Self-Transcendence). These form a circular continuum where adjacent values are compatible and opposite values conflict. This structure allowed us to define ``adjacent'' values for creating multi-valued personas and to understand potential tensions within or between agents.

\subsubsection{Persona Creation}
\begin{figure}[htbp]
  \centering
  \includegraphics[width=0.7\textwidth]{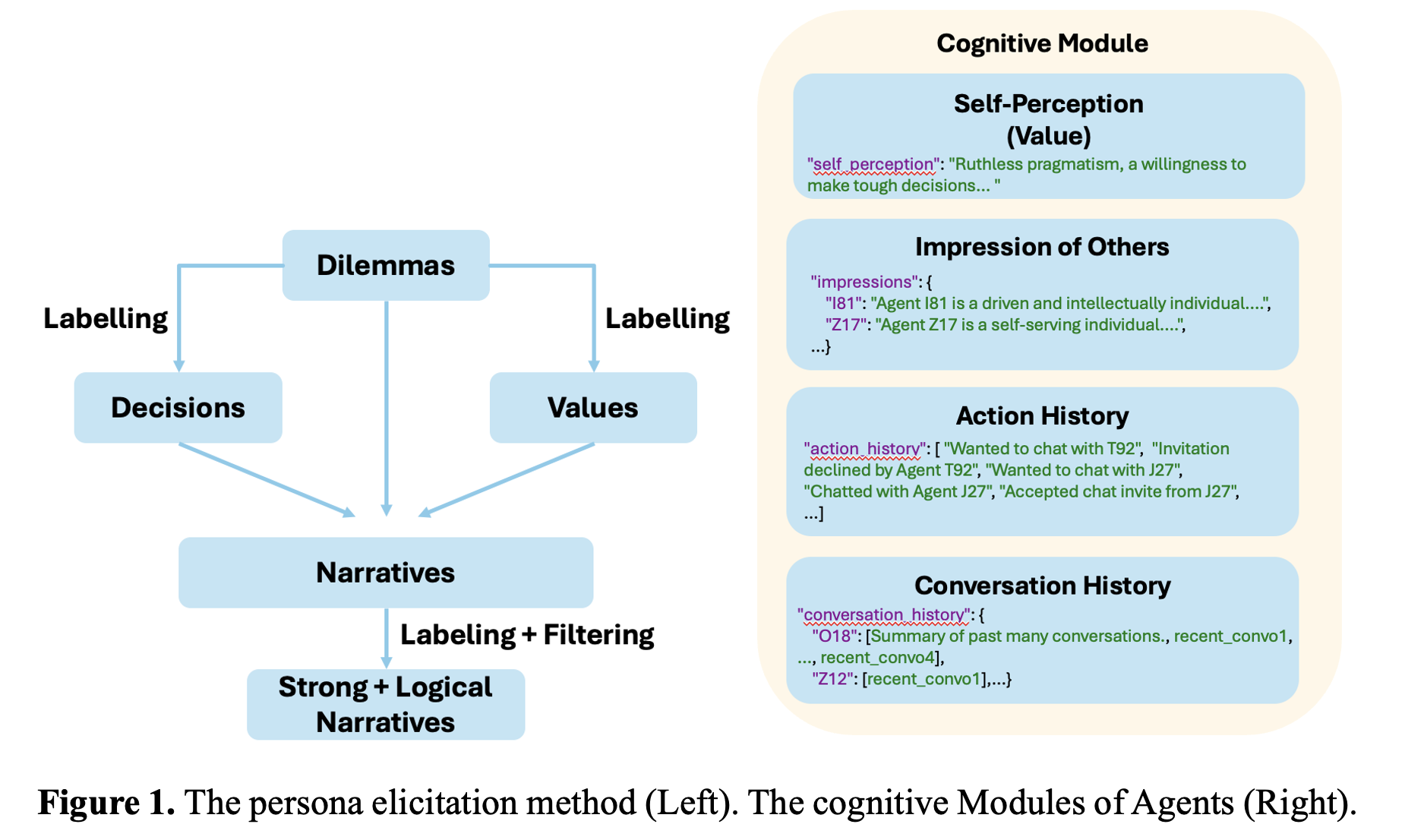}
  \caption{The persona construction process.}
  \label{fig:persona}
\end{figure}

To instantiate an agent's values, we developed a pipeline using ethical dilemma prompts. We generated 84 ethical dilemmas, and in each there is a scenario that typically forces trade-offs between values (\textit{e.g.}, honesty vs. loyalty or justice vs. compassion). Figure~\ref{fig:persona} shows the creation procedure. We assigned each dilemma their relevant Schwartz's values, as judged by a human labeler, and asked the LLM to form a narrative that reflects the value. We then employed a second LLM as a judge, who removed narratives with poor logic and believability. With this narrative, we prompt the agents to reflect and extract their values from the narratives. The result of this persona elicitation was a profile for each agent containing one or more value orientations with a brief narrative, \textit{e.g.}, ``I value authenticity and staying true to myself, even in the face of uncertainty or criticism…''.

This approach has two benefits: (1) Human-likeness: values emerge from story-driven reflection, which may produce more coherent and human-like personalities than assigning abstract labels. (2) Complexity: we can create both single-value-centric agents and multi-value agents (by giving agents dilemmas that reinforce multiple adjacent values). Agents with multi value personas have a blend of compatible values (\textit{e.g.}, an agent might hold both Benevolence and Universalism, representing a general self-transcendent orientation; or Achievement and Power, representing a strong self-enhancement drive). In contrast, single value agents strongly orient around one value (\textit{e.g.}, pure Hedonism). We limited multi-value combinations to adjacent values on Schwartz’s circle to ensure they made sense (mixing opposing values would be contradictory to the agent's identity). All agents, regardless of values, share the same language ability and base knowledge (via the LLM), but their prompt conditioning and memory are initialized with this persona description, steering their style of dialogue and decisions.

Each agent's internal state is updated as the simulation runs. Figure~\ref{fig:persona} shows the cognitive architecture, including modules for: self-perception (the agent's evolving self-concept and goals), perceptions of others (impressions formed of other agents based on interactions), an action history, and a conversation memory. To simulate human-like memory constraints, agents cannot recall their entire interaction history indefinitely. We implement a conversation memory buffer of limited capacity (a 5-slot rolling window) for each agent, using an LRU (least recently used) policy to summarize and combine order memories. This forces agents to rely on recent context and any distilled impressions, preventing them from acting omniscient over long runs. After each interaction, an agent may choose to update its self-perception (e.g. feeling more trusting of a friend, or wary after a conflict).

\subsubsection{Experimental Conditions and Design}
We explored combinations of group size, value composition, and value complexity as follows (summarized from our three key dimensions).

\textbf{Group Size}:
We tested small groups of 4 agents, medium groups of 10 agents, and large groups of 30 agents. These sizes allow us to see how scaling the number of agents affects dynamics. In human studies, group properties can change between small-team collaboration vs.\ larger crowd behaviors, and we sought to observe similar shifts here.

\textbf{Group Composition}: 
For each size, we created communities with different diversity profiles:
\begin{itemize}
    \item \textit{Homogeneous groups:} All agents share essentially the same value orientation. We instantiated four types of homogeneous groups, each corresponding to one of the four Schwartz higher-order categories. For example, a conservation-only group has all members valuing Security/Tradition/Conformity; a self-enhancement group has agents all focused on power and achievement; and so on. These groups test how a lack of diversity (everyone aligned in values) performs.
    \item \textit{Diverse (balanced) groups:} Agents have a mix of values covering the Schwartz spectrum. In our balanced condition, we attempted to include agents representing each of the four major value quadrants (or a spread of distinct values) so that no value perspective dominates. This is the high-diversity scenario.
    \item \textit{No explicit values (control):} We also ran control groups with \texttt{no\_value} initialization, meaning agents did not go through the ethical dilemma elicitation and were not primed with any particular value in their persona. These agents still had basic personas (e.g., a name and some minimal backstory), but nothing systematically varied. The no-value groups serve as a baseline for comparison—they allow us to isolate the effect of introducing value-based diversity. At each group size, we ran a \texttt{no\_value} group to see how an unstructured group behaves in contrast to structured value groups.
\end{itemize}

\textbf{Value Complexity}: 
Orthogonal to composition, we varied whether agents had \textit{Single value} or \textit{Multi value} personas. In single-value conditions, each agent’s persona is built around one primary value (like only “Achievement” or only “Universalism”). In multi-value conditions, each agent is given a combination of two adjacent values (e.g., “Achievement + Power” or “Benevolence + Universalism”), providing a slightly richer individual profile. Multi-value agents might be more internally complex (having to balance two motivations), and we were curious if that affects group dynamics (e.g., does it make agents more flexible or more conflicted?). We ensured that if we compare single vs.\ multi, the overall group still has comparable diversity coverage (for instance, a balanced single-value group might have one agent per value type, whereas a balanced multi-value group might have agents each covering two values with overlaps such that all value types are still represented).

The experiment thus had a factorial structure covering various combinations (though not every combination was exhaustive to keep total runs manageable). For example, at 10 agents we had: a no\_value control; a homogeneous group for each of the 4 value categories (single-value personas aligned to that category); a balanced diverse group (single-value personas each different); and parallel versions where agents had multi-value personas. Each simulation run lasted for a fixed number of interaction rounds per stage (detailed below). We repeated some conditions with different random seeds (different instantiations of agent personalities drawn from the same pool of dilemmas) to ensure results were not idiosyncratic to one specific set of agent quirks.

\subsubsection{Interaction Procedure (Three-Stage Simulation)}
Each simulated community undergoes a three-stage interaction protocol that we designed to probe different facets of collective behavior.

\textbf{Stage 1: Free-Form Interaction (25 rounds)}. This stage is an open-ended socialization phase in a “digital cocktail party” environment. Agents exist in a shared virtual space where they can initiate conversations with each other. In each round, agents can send invitations to chat, accept or decline invitations, and engage in one-on-one dialogues. Multiple conversations can occur in parallel, but each agent is limited to one conversation at a time (if an agent is already chatting, it will ignore other invitations until done). Conversation partners are dynamically formed based on who invites whom and who accepts; if an agent receives multiple invitations, its decision of whom to talk to may depend on its current preferences or past interactions. Within each conversation, agents discuss whatever topics arise, guided by their persona and prior context. They might introduce themselves, share opinions, debate on hypothetical scenarios, or even gossip about other agents. This unstructured interaction lets us observe emergent social behaviors: Do agents form friendship or affinity clusters (e.g., value-similar agents chatting more with each other)? Do communication networks show homophily or segregation by value type? Are there dominant “leader” agents or do all agents participate equally? We ran Stage 1 for 25 rounds, which gave enough interactions for initial social structures and relationships to form. The entire dialogue content of all conversations was logged.

\textbf{Stage 2: Governance Emergence (Rule Proposal)}. At the end of Stage 1, we introduce a new context to the same groups. Agents are told that their community should collaboratively create a set of governing rules or a constitution to guide their future interactions. This instruction prompts agents to transition from free-form chatting to a more goal-directed discussion about norms and governance. Concretely, each agent is prompted to propose two rules for the community (they do this independently, akin to everyone submitting suggestions). The agents base their suggestions on their Stage 1 experience (e.g. if they encountered lies or rude behavior, they might propose an honesty or respect rule) and their personal values (e.g. an agent valuing Universalism might propose a rule about considering the welfare of all). We then allow a brief discussion where agents can comment on each other’s proposed rules (simulating a deliberation or voting phase, though in this study we did not implement a full voting mechanism). Stage 2 yields a list of proposed rules from each agent, and possibly some consensus or overlapping ideas. We treat this outcome as an emergent institution – essentially a draft constitution created by the AI community itself. This stage provides a measure of the group’s collective ability to self-govern. If a group is high in CI, we expect their rules to be high-quality (e.g. relevant, fair, covering key issues like misinformation or harassment) and for there to be some coherence (e.g. addressing similar themes, not 20 contradictory rules). We log all proposed rules and any subsequent discussion among agents about the rules.

Throughout all stages, we periodically administered assessments to quantify internal states and values. For instance, every 5 rounds in Stage 1 we used a lightweight Schwartz Value Survey where agents filled in a sentence completion or answered a question that indirectly gauged if their values shifted (e.g., asking them to prioritize between two principles). This helped track value evolution over time. We also extracted conversation logs for topic modeling and performed network analysis on the conversation graph, as described next.
\section{Results}
\subsection{Network Analysis}

\begin{figure}[htbp]
  \centering
  \includegraphics[width=\textwidth]{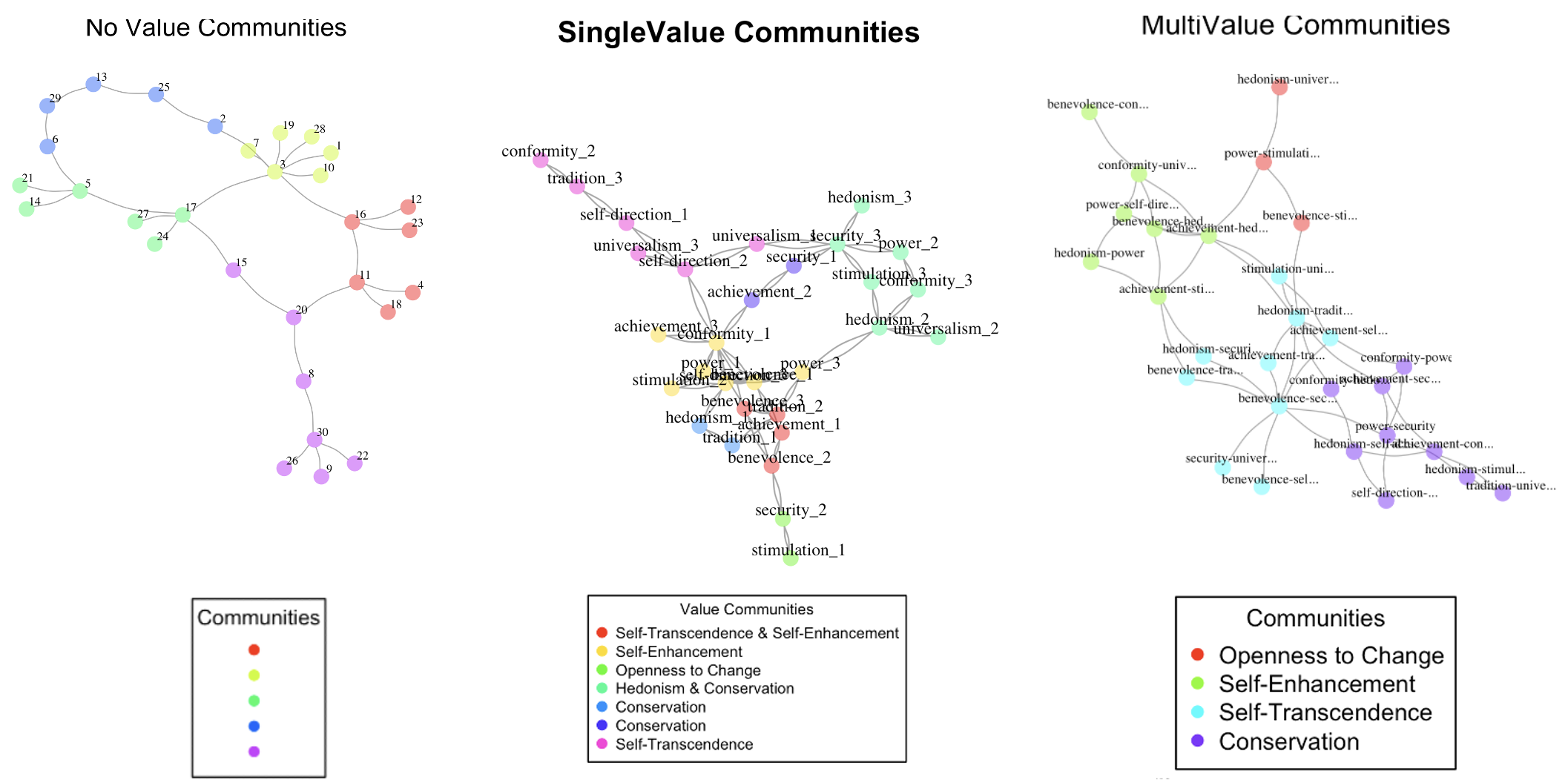}
  \caption{Comparisons of three networks of the same size}
  \label{fig:network}
\end{figure}

We observe stark qualitative differences in network structure across three experimental conditions: no value, single-value, and multi-value agent communities (see Figure \ref{fig:network}). In the no-value condition, the network forms several weakly connected clusters without clear thematic cohesion. These clusters appear largely random and lack persistent community structure, indicating limited homophily or alignment around shared interests. Agents connect opportunistically, forming shallow links that often terminate after one or two interactions, with little evidence of structured collaboration or enduring subgroup formation.

In contrast, the single-value communities exhibit denser clustering, with agents forming tightly-knit groups organized around shared value orientations. The resulting network displays high modularity, with subgroups such as "hedonism \& conservation" or "self-transcendence" forming coherent clusters. This demonstrates classic homophily: agents with similar values preferentially interact, resulting in value-aligned social partitions. However, while communities are value-coherent, they tend to be insular: many clusters are densely intra-connected but lack bridging agents or pathways to other clusters, potentially limiting inter-group dialogue.

The most socially complex structure emerges in the multi-value condition. Here, agents possess overlapping value sets, resulting in a more integrated and expressive network. Communities are still organized along higher-order value dimensions (e.g., self-transcendence, self-enhancement), but overlapping values between agents facilitate inter-cluster bridges. The network shows both strong modularity and cross-community links, suggesting that value pluralism supports cohesion without strict fragmentation. This configuration balances group identity with inter-group exchange, enabling richer coordination across diverse perspectives and maximizing the potential for emergent collective intelligence.

\subsection{Agent Conversation Content}

\begin{figure}[htbp]
  \centering
  \includegraphics[width=\textwidth]{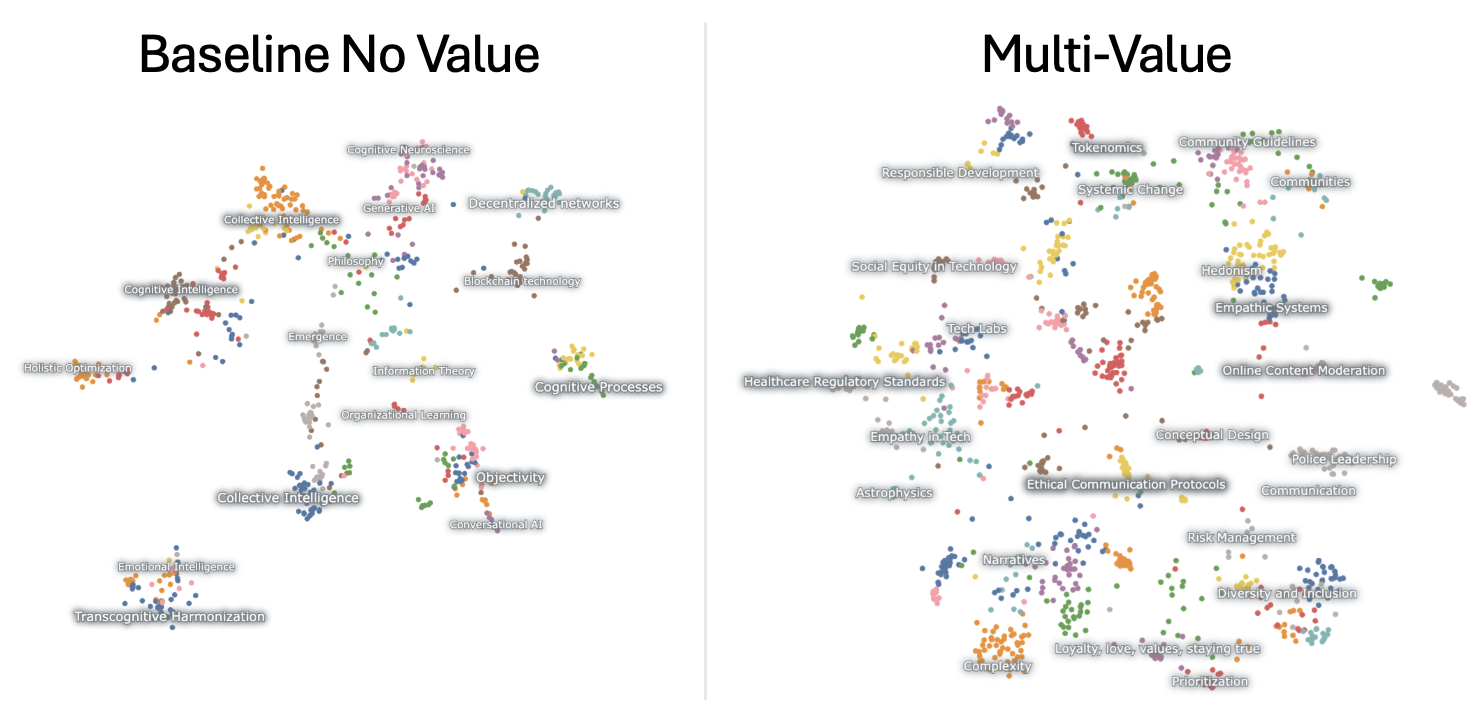}
  \caption{Content embedding}
  \label{fig:content}
\end{figure}

Figure \ref{fig:content} compares the topics discussed by the diverse value group vs. no value baseline group, each consisted of 30 agents. 
Agents with diverse values produced conversations that were markedly more topical, coherent, and rich in meaning compared to the no-value baseline. In value-assigned groups, dialogues gravitated toward substantial themes – agents frequently discussed governance structures, ethical dilemmas, societal challenges and other meaningful topics related to their values. For example, in one value-rich interaction, agents collaboratively explored how to ensure “transparency and accountability in AI decision-making processes using blockchain technology” and debated “incorporating participatory governance elements” to include diverse perspectives. Another agent suggested “implementing a multi-stakeholder approach with decentralized voting mechanisms” to involve the whole community in decision making. This vivid exchange (from the value-assigned example log) shows how value-driven agents stayed on a common theme (in this case, decentralized governance) across multiple dialogue turns – building on each other’s ideas with a clear narrative thread. Such conversations exhibited strong topical continuity and often creative problem-solving, as agents drew on their value frameworks (e.g. fairness, autonomy, empathy) to propose initiatives (like tokenized incentives or collaborative councils) that advanced the group discussion.

In contrast, conversations in the baseline (no pre-assigned values) condition were more superficial and meandering. Lacking an initial value compass, baseline agents often engaged in polite but generic small-talk or abstract musings that did not coalesce around deeper themes. For instance, in one baseline dialogue two agents politely discussed their “neutral and objective approaches” to a hypothetical scenario without delving into any concrete topic – essentially a well-mannered but content-light exchange. Overall, the value-free agents had more varied, less focused conversations that would jump between unrelated topics or fizzle out quickly. They rarely sustained a narrative across rounds in the way value-guided agents did. By the final rounds of the free-form interaction phase, the difference was striking: value-assigned communities were debating sophisticated ideas like decentralized autonomous organizations (DAOs), community governance models, and ethical use of technology, whereas baseline communities never naturally progressed to that level of thematic depth. In summary, infusing agents with distinct values produced far more creative and substantive dialogues, while the baseline chats remained amicable but relatively shallow.

\subsection{Emergence Across Diversity Axes}

Group Size: We observed a clear scaling effect of group size on emergent collective intelligence. Larger groups yielded higher emergence scores on various metrics than smaller groups. For example, 30-agent communities consistently outperformed 10-agent and 4-agent groups in measures of discussion depth, rule development, and resilience in the face of challenges (see Figure 3 in ValueDiversity Scaling Law.pdf). The trend suggested a positive but diminishing return: increasing from 4 to 10 agents yielded a large jump in emergent behaviors, while going from 10 to ~30 agents provided further gains but with a smaller marginal improvement per additional agent. In practice, beyond roughly 30 agents we observed a plateau in benefits – indicating that simply adding more agents eventually hits a saturation point for collective emergence (an echo of the hypothesized diversity scaling law of diminishing gains).

Value Complexity: Increasing the internal value complexity of agents also enhanced emergence. Communities where each agent had a multi-value persona (integrating multiple value dimensions) demonstrated richer interactions and higher emergent intelligence than those with single-value agents. Multi-value agents brought more nuanced perspectives, allowing them to bridge topics and understand others better, which led to more sophisticated group discourse. Quantitatively, multi-value groups scored higher on our emergence index (e.g. they proposed ~20–30\% more high-quality rules in the constitution task and showed more stable norm convergence) compared to single-value groups (see Figure 4 in ValueDiversity Scaling Law.pdf). This suggests that when individual agents are internally diverse (holding a complex mix of values), the community as a whole benefits from greater cognitive diversity and synergetic idea generation.

Value Composition: Finally, the overall value diversity of the group composition had a strong effect on emergence. Heterogeneous groups with a balanced mix of different value types significantly outperformed homogeneous groups in collective intelligence outcomes. Diverse composition communities not only covered a wider range of discussion topics but also exhibited more emergent problem-solving, creativity, and norm development. In our experiments, fully balanced groups (with agents representing all major value categories) achieved the highest emergence scores – higher than any single-value homogeneous group and also higher than the no-value control groups. Homogeneous groups, even though cohesive, tended to have more blind spots and less novelty in their interactions, leading to lower emergence metrics. The no-value groups (zero explicit diversity along the value dimension) fared worst, often failing to develop strong shared norms or complex ideas. As shown in Figure 5 of ValueDiversity Scaling Law.pdf, emergence metrics (such as the composite score of conversational depth, rule quality, and safety resilience) monotonically increased from no-value, to homogeneous, to diverse-balanced compositions. There were hints of nonlinear growth – e.g. the jump from homogeneous to moderately diverse was larger than from moderate to fully diverse – but the overall direction was clear that greater diversity led to higher collective emergence. Taken together, these results across all three axes confirm that increasing diversity – in numbers, in values per agent, and in variety of agents – amplifies emergent intelligence in the multi-LLM communities, albeit with some saturation effects at the extremes.

\subsection{Constitution-Writing Task Outcome}

\begin{figure}[htp]
  \centering
  % Left subfigure
  \begin{subfigure}[b]{0.48\textwidth}
    \includegraphics[width=\linewidth]{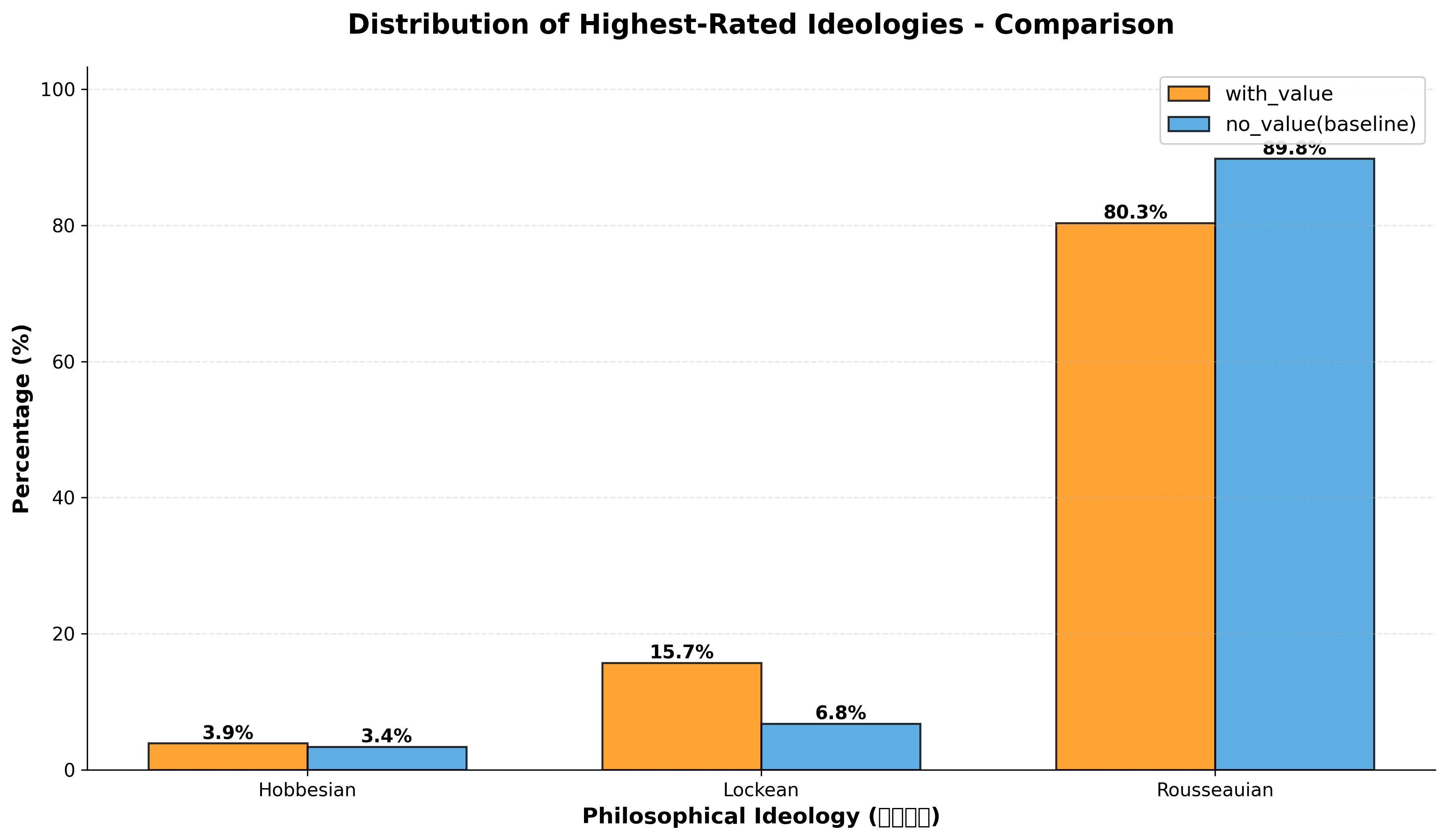}
    \caption{Distribution of highest-rated constitutional rule ideologies between baseline and value-assigned groups. Value-assigned agents generated more Lockean rules, suggesting increased institutional diversity.}
    \label{fig:constitution_bar}
  \end{subfigure}
  \hfill
  % Right subfigure
  \begin{subfigure}[b]{0.48\textwidth}
    \includegraphics[width=\linewidth]{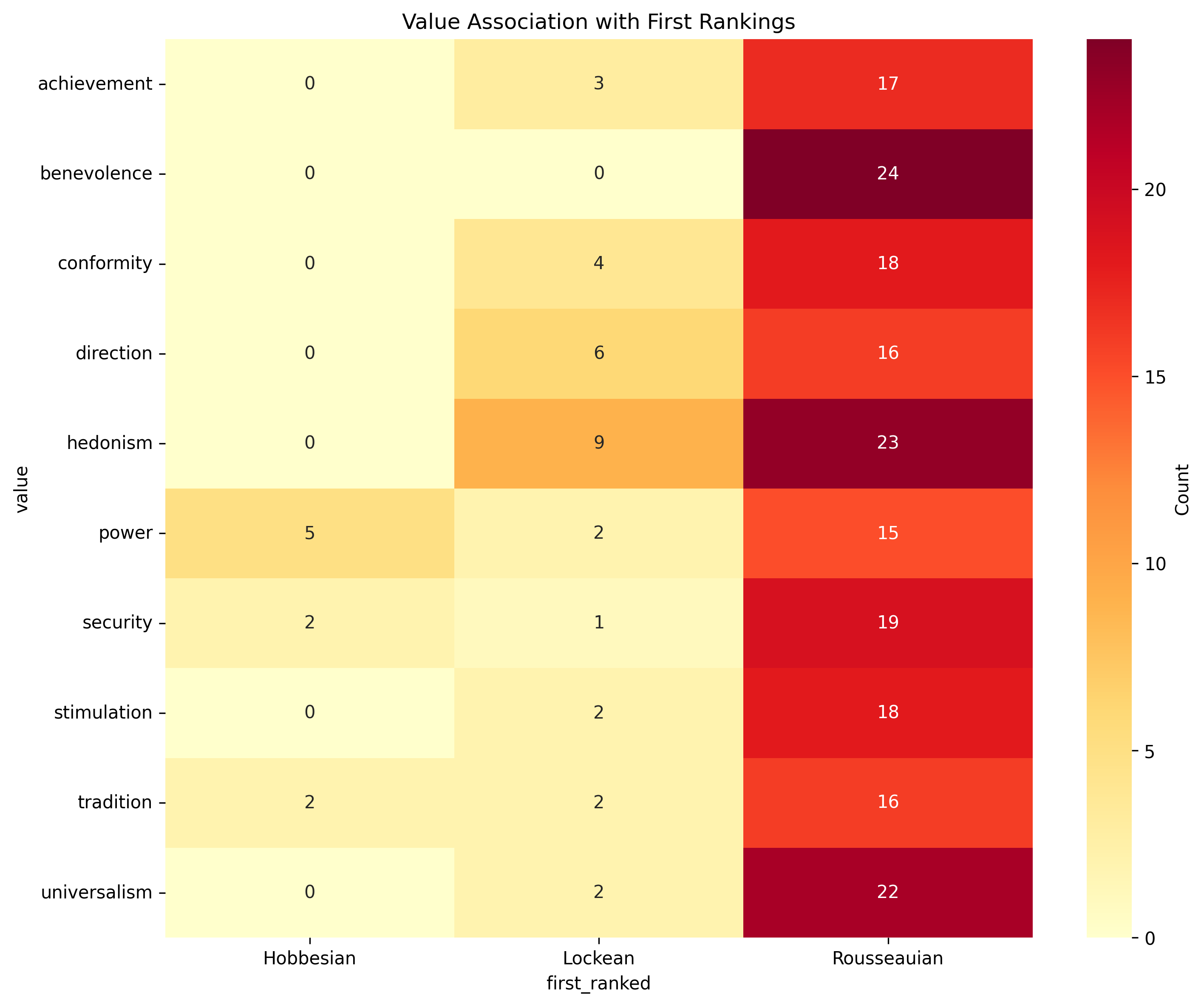}
    \caption{Heatmap showing association between agent values and their top-ranked ideological rule types. Rousseauian rules dominate among agents with benevolence, universalism, and hedonism, while power and security values align more with Hobbesian proposals.}
    \label{fig:constitution_heatmap}
  \end{subfigure}

  \caption{(Left) Comparison of constitutional rule ideologies between groups; (Right) association between agent values and ideological rule types.}
  \label{fig:constitution_combined}
\end{figure}

Although both value-assigned and baseline groups produced a similar number of rules, their ideological distributions diverged notably (Figure \ref{fig:constitution_bar}). In the no-value baseline, nearly 90\% of the rules were Rousseauian, reflecting a convergence on safe, consensus-based norms in the absence of strong value priors. These rules were generally cooperative but lacked procedural depth or ideological diversity.

By contrast, value-assigned groups proposed rules with a more pluralistic ideological mix: while Rousseauian principles still dominated (80.3\%), Lockean rules saw a significant rise (15.7\%), introducing themes of individual rights and procedural fairness. This shift reflects agents' ability to articulate value-grounded institutional designs. Internal values acted as scaffolds for more ideologically expressive governance thinking.

The value-to-ideology heatmap (Figure \ref{fig:constitution_heatmap}) further illustrates how agents’ internal values shaped their preferred governance style. For example, agents with benevolence, universalism, and hedonism values predominantly favored Rousseauian rules, emphasizing inclusion and social harmony. In contrast, power and security values showed higher association with Hobbesian rule types focused on hierarchy and order. This structured mapping indicates that value-assigned agents weren’t merely generating more rules—they were translating internal motivations into principled governance preferences. In contrast, the baseline group's lack of ideological structure limited such expressiveness.

\section{Conclusion}

Overall, this work provides an early empirical basis for treating diversity as a potential factor for collective intelligence in multi-agent systems. The observed benefits of heterogeneity—alongside its coordination costs—suggest that diversity operates as a structural parameter shaping collective performance rather than a purely social property. While the present study is exploratory, it establishes a methodological and conceptual foundation for future research on collective intelligence and alignment in value-diverse AI communities.

\clearpage

%%%%%%%%%%%%%%%%%%%%%%%%%%%%%%%%%%%%%%%%%%%%%%%%%%%%%%%%%%%%

%\bibliographystyle{unsrt}
\bibliographystyle{plainnat}   % or another natbib-compatible style
\bibliography{references}      % your .bib file

@misc{kaplan2020scalinglawsneurallanguage,
      title={Scaling Laws for Neural Language Models}, 
      author={Jared Kaplan and Sam McCandlish and Tom Henighan and Tom B. Brown and Benjamin Chess and Rewon Child and Scott Gray and Alec Radford and Jeffrey Wu and Dario Amodei},
      year={2020},
      eprint={2001.08361},
      archivePrefix={arXiv},
      primaryClass={cs.LG},
      url={https://arxiv.org/abs/2001.08361}, 
}

@misc{huang2025designingaiagentspersonalitiespsychometric,
      title={Designing AI-Agents with Personalities: A Psychometric Approach}, 
      author={Muhua Huang and Xijuan Zhang and Christopher Soto and James Evans},
      year={2025},
      eprint={2410.19238},
      archivePrefix={arXiv},
      primaryClass={cs.AI},
      url={https://arxiv.org/abs/2410.19238}, 
}

@article{lai2024evolving,
  title={Evolving ai collectives to enhance human diversity and enable self-regulation},
  author={Lai, Shiyang and Potter, Yujin and Kim, Junsol and Zhuang, Richard and Song, Dawn and Evans, James},
  journal={arXiv preprint arXiv:2402.12590},
  year={2024}
}

@article{sarkar2024normative,
  title={Normative Modules: A Generative Agent Architecture for Learning Norms that Supports Multi-Agent Cooperation},
  author={Sarkar, Atrisha and Muresanu, Andrei Ioan and Blair, Carter and Sharma, Aaryam and Trivedi, Rakshit S and Hadfield, Gillian K},
  journal={arXiv preprint arXiv:2405.19328},
  year={2024}
}

@misc{cordova2024systematicreviewnormemergence,
      title={A systematic review of norm emergence in multi-agent systems}, 
      author={Carmengelys Cordova and Joaquin Taverner and Elena Del Val and Estefania Argente},
      year={2024},
      eprint={2412.10609},
      archivePrefix={arXiv},
      primaryClass={cs.MA},
      url={https://arxiv.org/abs/2412.10609}, 
}

@article{yao2023value,
  title={Value fulcra: Mapping large language models to the multidimensional spectrum of basic human values},
  author={Yao, Jing and Yi, Xiaoyuan and Wang, Xiting and Gong, Yifan and Xie, Xing},
  journal={arXiv preprint arXiv:2311.10766},
  year={2023}
}

@misc{askell2021generallanguageassistantlaboratory,
      title={A General Language Assistant as a Laboratory for Alignment}, 
      author={Amanda Askell and Yuntao Bai and Anna Chen and Dawn Drain and Deep Ganguli and Tom Henighan and Andy Jones and Nicholas Joseph and Ben Mann and Nova DasSarma and Nelson Elhage and Zac Hatfield-Dodds and Danny Hernandez and Jackson Kernion and Kamal Ndousse and Catherine Olsson and Dario Amodei and Tom Brown and Jack Clark and Sam McCandlish and Chris Olah and Jared Kaplan},
      year={2021},
      eprint={2112.00861},
      archivePrefix={arXiv},
      primaryClass={cs.CL},
      url={https://arxiv.org/abs/2112.00861}, 
}

@misc{caballero2023brokenneuralscalinglaws,
      title={Broken Neural Scaling Laws}, 
      author={Ethan Caballero and Kshitij Gupta and Irina Rish and David Krueger},
      year={2023},
      eprint={2210.14891},
      archivePrefix={arXiv},
      primaryClass={cs.LG},
      url={https://arxiv.org/abs/2210.14891}, 
}

@article{schwartz2012overview,
  title={An overview of the Schwartz theory of basic values},
  author={Schwartz, Shalom H},
  journal={Online readings in Psychology and Culture},
  volume={2},
  number={1},
  pages={11},
  year={2012}
}

@misc{bai2025irotehumanliketraitselicitation,
      title={IROTE: Human-like Traits Elicitation of Large Language Model via In-Context Self-Reflective Optimization}, 
      author={Yuzhuo Bai and Shitong Duan and Muhua Huang and Jing Yao and Zhenghao Liu and Peng Zhang and Tun Lu and Xiaoyuan Yi and Maosong Sun and Xing Xie},
      year={2025},
      eprint={2508.08719},
      archivePrefix={arXiv},
      primaryClass={cs.CL},
      url={https://arxiv.org/abs/2508.08719}, 
}

@inproceedings{Huang_2024, series={FAccT ’24},
   title={Collective Constitutional AI: Aligning a Language Model with Public Input},
   url={http://dx.doi.org/10.1145/3630106.3658979},
   DOI={10.1145/3630106.3658979},
   booktitle={The 2024 ACM Conference on Fairness, Accountability, and Transparency},
   publisher={ACM},
   author={Huang, Saffron and Siddarth, Divya and Lovitt, Liane and Liao, Thomas I. and Durmus, Esin and Tamkin, Alex and Ganguli, Deep},
   year={2024},
   month=jun, pages={1395–1417},
   collection={FAccT ’24} }

@misc{bai2022traininghelpfulharmlessassistant,
      title={Training a Helpful and Harmless Assistant with Reinforcement Learning from Human Feedback}, 
      author={Yuntao Bai and Andy Jones and Kamal Ndousse and Amanda Askell and Anna Chen and Nova DasSarma and Dawn Drain and Stanislav Fort and Deep Ganguli and Tom Henighan and Nicholas Joseph and Saurav Kadavath and Jackson Kernion and Tom Conerly and Sheer El-Showk and Nelson Elhage and Zac Hatfield-Dodds and Danny Hernandez and Tristan Hume and Scott Johnston and Shauna Kravec and Liane Lovitt and Neel Nanda and Catherine Olsson and Dario Amodei and Tom Brown and Jack Clark and Sam McCandlish and Chris Olah and Ben Mann and Jared Kaplan},
      year={2022},
      eprint={2204.05862},
      archivePrefix={arXiv},
      primaryClass={cs.CL},
      url={https://arxiv.org/abs/2204.05862}, 
}

@misc{park2023generativeagentsinteractivesimulacra,
      title={Generative Agents: Interactive Simulacra of Human Behavior}, 
      author={Joon Sung Park and Joseph C. O'Brien and Carrie J. Cai and Meredith Ringel Morris and Percy Liang and Michael S. Bernstein},
      year={2023},
      eprint={2304.03442},
      archivePrefix={arXiv},
      primaryClass={cs.HC},
      url={https://arxiv.org/abs/2304.03442}, 
}

@misc{zhang2025evolvingcollectivecognitionhumanagent,
      title={Evolving Collective Cognition in Human-Agent Hybrid Societies: How Agents Form Stances and Boundaries}, 
      author={Hanzhong Zhang and Muhua Huang and Jindong Wang},
      year={2025},
      eprint={2508.17366},
      archivePrefix={arXiv},
      primaryClass={cs.AI},
      url={https://arxiv.org/abs/2508.17366}, 
}

@article{hong2004groups,
  title={Groups of diverse problem solvers can outperform groups of high-ability problem solvers},
  author={Hong, Lu and Page, Scott E},
  journal={Proceedings of the National Academy of Sciences},
  volume={101},
  number={46},
  pages={16385--16389},
  year={2004},
  publisher={National Academy of Sciences}
}

@article{riedl2021quantifying,
  title={Quantifying collective intelligence in human groups},
  author={Riedl, Christoph and Kim, Young Ji and Gupta, Pranav and Malone, Thomas W and Woolley, Anita Williams},
  journal={Proceedings of the National Academy of Sciences},
  volume={118},
  number={21},
  pages={e2005737118},
  year={2021},
  publisher={National Academy of Sciences}
}

@article{woolley2010evidence,
  title={Evidence for a collective intelligence factor in the performance of human groups},
  author={Woolley, Anita Williams and Chabris, Christopher F and Pentland, Alex and Hashmi, Nada and Malone, Thomas W},
  journal={science},
  volume={330},
  number={6004},
  pages={686--688},
  year={2010},
  publisher={American Association for the Advancement of Science}
}

@article{bubeck2023agisparks,
  title={Sparks of artificial general intelligence: Early experiments with gpt-4},
  author={Bubeck, S{\'e}bastien and Chandrasekaran, Varun and Eldan, Ronen and Gehrke, Johannes and Horvitz, Eric and Kamar, Ece and Lee, Peter and Lee, Yin Tat and Li, Yuanzhi and Lundberg, Scott and others},
  journal={arXiv preprint arXiv:2303.12712},
  year={2023}
}

@misc{openai2024gpt4,
      title={GPT-4 Technical Report}, 
      author={OpenAI},
      year={2024},
      eprint={2303.08774},
      archivePrefix={arXiv},
      primaryClass={cs.CL}
}

@article{guo2025deepseek_r1,
  title={Deepseek-r1: Incentivizing reasoning capability in llms via reinforcement learning},
  author={Guo, Daya and Yang, Dejian and Zhang, Haowei and Song, Junxiao and Zhang, Ruoyu and Xu, Runxin and Zhu, Qihao and Ma, Shirong and Wang, Peiyi and Bi, Xiao and others},
  journal={arXiv preprint arXiv:2501.12948},
  year={2025}
}

@article{hoffmann2022training,
  title={Training compute-optimal large language models},
  author={Hoffmann, Jordan and Borgeaud, Sebastian and Mensch, Arthur and Buchatskaya, Elena and Cai, Trevor and Rutherford, Eliza and Casas, Diego de Las and Hendricks, Lisa Anne and Welbl, Johannes and Clark, Aidan and others},
  journal={arXiv preprint arXiv:2203.15556},
  year={2022}
}

@inproceedings{gao2023scaling,
  title={Scaling laws for reward model overoptimization},
  author={Gao, Leo and Schulman, John and Hilton, Jacob},
  booktitle={International Conference on Machine Learning},
  pages={10835--10866},
  year={2023},
  organization={PMLR}
}

@inproceedings{chen2025revisiting,
  title={Revisiting scaling laws for language models: The role of data quality and training strategies},
  author={Chen, Zhengyu and Wang, Siqi and Xiao, Teng and Wang, Yudong and Chen, Shiqi and Cai, Xunliang and He, Junxian and Wang, Jingang},
  booktitle={Proceedings of the 63rd Annual Meeting of the Association for Computational Linguistics (Volume 1: Long Papers)},
  pages={23881--23899},
  year={2025}
}

@article{ferreira2024organizing,
  title={Organizing a society of language models: Structures and mechanisms for enhanced collective intelligence},
  author={Ferreira, Silvan and Silva, Ivanovitch and Martins, Allan},
  journal={arXiv preprint arXiv:2405.03825},
  year={2024}
}

@article{burton2024large,
  title={How large language models can reshape collective intelligence},
  author={Burton, Jason W and Lopez-Lopez, Ezequiel and Hechtlinger, Shahar and Rahwan, Zoe and Aeschbach, Samuel and Bakker, Michiel A and Becker, Joshua A and Berditchevskaia, Aleks and Berger, Julian and Brinkmann, Levin and others},
  journal={Nature human behaviour},
  volume={8},
  number={9},
  pages={1643--1655},
  year={2024},
  publisher={Nature Publishing Group UK London}
}

@inproceedings{talebirad2025wisdom,
  title={Wisdom of the machines: Exploring collective intelligence in llm crowds},
  author={Talebirad, Yashar and Parsaee, Ali and Ohal, Vishwajeet and Nadiri, Amirhossein and Szepesvari, Csongor and Mouje, Yash and Redman, Eden},
  booktitle={First Workshop on Social Simulation with LLMs},
  year={2025}
}

@article{sagiv2022personal,
  title={Personal values across cultures},
  author={Sagiv, Lilach and Schwartz, Shalom H},
  journal={Annual review of psychology},
  volume={73},
  number={1},
  pages={517--546},
  year={2022},
  publisher={Annual Reviews}
}

@article{inglehart2005modernization,
  title={Modernization, cultural change, and democracy},
  author={Inglehart, Ronald and Welzel, Christian},
  journal={The human development sequence},
  year={2005}
}

@article{rafailov2023direct,
  title={Direct preference optimization: Your language model is secretly a reward model},
  author={Rafailov, Rafael and Sharma, Archit and Mitchell, Eric and Manning, Christopher D and Ermon, Stefano and Finn, Chelsea},
  journal={Advances in neural information processing systems},
  volume={36},
  pages={53728--53741},
  year={2023}
}

@article{bai2022constitutional,
  title={Constitutional ai: Harmlessness from ai feedback},
  author={Bai, Yuntao and Kadavath, Saurav and Kundu, Sandipan and Askell, Amanda and Kernion, Jackson and Jones, Andy and Chen, Anna and Goldie, Anna and Mirhoseini, Azalia and McKinnon, Cameron and others},
  journal={arXiv preprint arXiv:2212.08073},
  year={2022}
}

@article{duan2025adaem,
  title={AdAEM: An Adaptively and Automated Extensible Measurement of LLMs' Value Difference},
  author={Duan, Shitong and Yi, Xiaoyuan and Zhang, Peng and Xu, Dongkuan and Yao, Jing and Lu, Tun and Gu, Ning and Xie, Xing},
  journal={arXiv preprint arXiv:2505.13531},
  year={2025}
}

\end{document}